\documentclass[]{style/ceurart}
\usepackage{listings}
\usepackage{array}
\begin{document}

\copyrightyear{2024}
\copyrightclause{Copyright for this paper by its authors.
  Use permitted under Creative Commons License Attribution 4.0
  International (CC BY 4.0).}

\conference{CLEF 2024: Conference and Labs of the Evaluation Forum, September 09-12, 2024, Grenoble, France}

\title{Biomedical Nested NER with Large Language Model and UMLS Heuristics}
\subtitle{DS@GT CLEF2024 BioNNE Competition Working Note}
\author[1]{Wenxin Zhou}[
orcid=0009-0002-3325-3357,
email=wzhou77@gatech.edu,
]
\cormark[1]

\address[1]{Georgia Institute of Technology, North Ave NW, Atlanta, GA 30332, United States}
\cortext[1]{Corresponding author.}

\begin{abstract}
  In this paper, we present our system for the BioNNE English track, which aims to extract 8 types of biomedical nested named entities from biomedical text. We use a large language model (Mixtral 8x7B instruct) and ScispaCy NER model to identify entities in an article and build custom heuristics based on unified medical language system (UMLS) semantic types to categorize the entities. We discuss the results and limitations of our system and propose future improvements. Our system achieved an F1 score of 0.39 on the BioNNE validation set and 0.348 on the test set.
\end{abstract}

\begin{keywords}
  large language model \sep
  prompt engineering \sep
  named entity recognition \sep
  nested named entity recognition
\end{keywords}

\maketitle

\section{Introduction}
Named entity recognition is an information extraction task that aims to identify named entities in text and classify them into predefined categories.
Nested named entity recognition involves detecting both the outer entities and the inner entities.
The BioNNE competition \cite{bionne}, being part of CLEF 2024 BioASQ lab \cite{BioASQ2024overview}, focuses on extracting nested named entities from biomedical text.
The nested named entity types include anatomy (ANATOMY), chemicals (CHEM), diseases (DISO), physiology (PHYS), scientific findings (FINDING), injury or poisoning damages (INJURY\_POISONING),
lab procedures (LABPROC), and medical devices (DEVICE) \cite{nerelbio}.

The challenge offers Russian, English and bilingual tracks. For the English track, the organizers provided a training set with 50 records and a validation set with 50 records.
Each record contains a text, which is a PubMed abstract, and a list of entities annotated in BRAT format with the starting and ending locations of the entities in the text.
In the testing phase, the organizers released a test set with 154 abstracts and 346 extra files, which resulted in 500 records in total.

Our team focuses on the BioNNE English track.
Our system uses a large language model (specifically Mixtral 8x7B instruct model \cite{Jiang2024MixtralOE}) and a biomedical NER model to find entities in an article.
Then, the system uses unified medical language system (UMLS) semantic types to filter and aggregate the entities.
The implementation can be found on Github\footnote{https://github.com/dsgt-kaggle-clef/bioasq-2024}.

\section{Related Work}
\subsection{Nested NER}
The state-of-the-art nested NER models are the machine reading comprehension (MRC) model \cite{stoamrc} and the sequence learning model \cite{stoaseq}.
Loukachevitch et al. trained the MRC and sequence model on the NEREL-BIO dataset \cite{nerelbio}, which is the predecessor of the BioNNE dataset.
The two datasets are based on PubMed abstracts and with the nested NER annotations. The difference is that the NEREL-BIO dataset contains more entity types, including the BioNNE entity types and additional entity types such as food, gene and activity.
Loukachevitch et al. were able to achieve a macro-F1 score of 0.5968 with MRC model on the NEREL-BIO dataset. 

\subsection{Large Language Model}
Recently, large language models (LLM) have shown great potential in solving NLP tasks. Since they are pre-trained on a large corpus,
they can solve various problems in different domains including biomedical question answering and information extraction.
Prompt engineering \cite{promptengineering} is one of the key techniques to interact with LLM. 
The model is given a prompt with instructions and examples. The model can then generate the response according to the prompt.
This is also known as LLM few-shot in-context learning. The quality of the prompt affects model output and performance.
As a result, a lot of research has been done on prompt engineering to optimize the performance of LLM on specific tasks,
as an alternative to model finetuning, because finetuning is more expensive and time-consuming.

Chen et al. \cite{chatgptBLURB} measured the performance of GPT-3.5 on the Biomedical Language Understanding and Reasoning Benchmark (BLURB).
GPT-3.5 achieved 58.5 while the SOTA model had a score of 84.5. 
Although the existing LLM models are not as good as the SOTA biomedical models, LLMs show great potential in text reasoning and generation.
At the same time, LLMs have limitations such as hallucination and inconsistency \cite{llmLimitation}.

Mixtral 8x7B \cite{Jiang2024MixtralOE}, a sparse mixture of experts model (SMoE) is one of the well-known LLM models,
which can process five languages including English, French, Italian, German and Spanish with a context of 32k tokens.
The basic model and the instruct model of Mixtral 8x7B achieve better performance than the counterparts of GPT-3.5 \cite{chatgpt35} and Llama2 70B \cite{llama2} in several benchmarks.
Mixtral 8x7B only uses 13B parameters during inference time, which makes it more efficient than other models with the same performance.
Therefore, we use Mixtral 8x7B instruct as our LLM model in this study.

\section{Methodology}

\begin{figure}
  \centering
  \includegraphics[width=\linewidth]{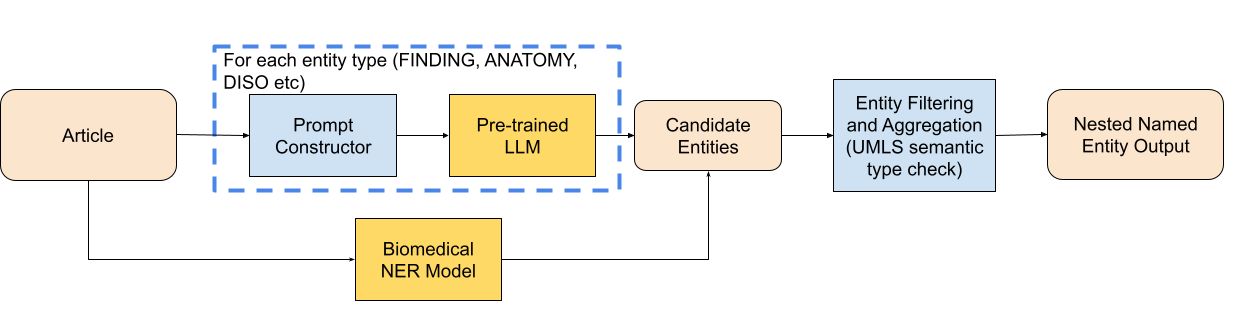}
  \caption{BioNNE System Design}
  \label{fig:design}
\end{figure}

\begin{figure}
  \centering
  \includegraphics[width=\linewidth]{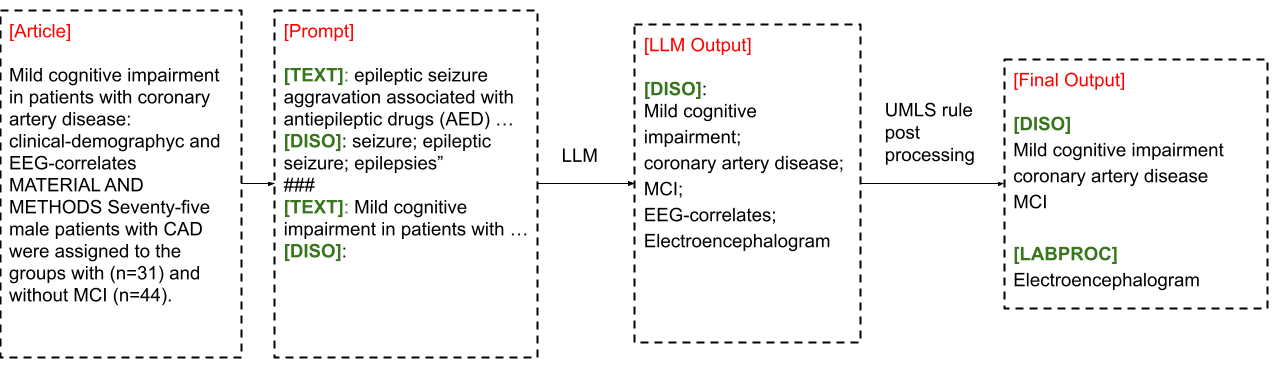}
  \caption{BioNNE Data Flow}
  \label{fig:data_flow}
\end{figure}

We build a system for the BioNNE English track that uses a general-purpose LLM and a biomedical domain language model to identify entities in an article 
and then uses custom heuristics based on unified medical language system (UMLS) semantic types to determine the entity types. 
The system design is shown in Figure \ref{fig:design}. 
Figure \ref{fig:data_flow} shows an example of how DISO and LABPROC entities are extracted from an article by LLM and categorized by UMLS heuristics.


\subsection{LLM and Prompt Engineering}
Given an article abstract, we use Mixtral 8x7B instruct LLM model to find the entities for each category.
We construct a prompt for each category using two of the examples in the training dataset and an instruction that describes the entity type.
Then the LLM returns a list of entities separated by semicolon in the response. We then parse the returned entities and store them as candidates for the category.
We perform post-processing on the response list to remove duplicates as well as entities that are not present in the original text.
If the entities in the returned string are not separated by semicolons, we resample the LLM output.
For example, Figure \ref{fig:promptexp} shows the prompt and response to query ANATOMY entity for record 26271422\_en.

\begin{figure}[h!]
  \centering
  \begin{flushleft}
    \textbf{Prompt:}
  \end{flushleft}
  
\fbox{
\begin{minipage}{0.9\linewidth}
Instruction: Return phrases or entities that comprise organs, body part, cells and cell components, body substances in TEXT, in the ANATOMY concatenated by ";"\\

\lbrack TEXT\rbrack: Impact of bosentan therapy on stress-induced pulmonary hypertension in patients with systemic sclerosis. AIM To describe hemodynamic and clinical changes in patients with elevated mean pulmonary artery pressure (MPAP) ...\\
\lbrack ANATOMY\rbrack: pulmonary; artery; pulmonary artery; lung;heart; left heart; atrial; right atrial; cardiac; arterial; vascular; pulmonary arterial; pulmonary vascular\\
\#\#\#\\
\lbrack TEXT\rbrack: The authors present the material of their study of the morphological and molecular biological features of damage to the stem cell ...\\
\lbrack ANATOMY\rbrack: lung biopsies; respiratory acinus; lung tissue; mesenchymal cell; myofibroblast; mesenchymal stem cell; SCN; stem cell; cell; lung;pulmonary; acinus; stem cell niches; tissue;mesenchymal; SCN areas; respiratory acini; biopsies; sections; acini; cells\\
\#\#\#\\
\lbrack TEXT\rbrack: Endoprosthetic replacement with lifting of abdominal wall in treatment of umbilical and postoperative ventral hernias. The results of ...\\
\lbrack ANATOMY\rbrack:\\
\end{minipage}}
\break

\vspace{10pt} 
\begin{flushleft}
  \textbf{Response:}
\end{flushleft}

\fbox{
\begin{minipage}{0.9\linewidth}
abdominal wall defect; abdomen ptosis; polypropylene prosthesis; main flap; roundish edges; additional flap; super lightweight; hypogastric area; musculoaponeurotic tissues lifting; physical health component; psychic health component; standard; original
\end{minipage}}
\caption{Example Prompt and Response for ANATOMY entity}
\label{fig:promptexp}
\end{figure}

\subsection{Biomedical NER model}
In addition to LLM, we also use the ScispaCy biomedical NER model {\it en\_ner\_bc5cdr\_md} \cite{scispacy_ner} to find the named entities. 
The ScispaCy model is trained on the BC5CDR corpus, which detects DISEASE and CHEMICAL entities in the document.
We store the entities found by the ScispaCy model as candidates for the DISO and CHEM categories.

\subsection{UMLS Heuristics}
Finally, we query UMLS \cite{umls_paper} to find the semantic types for the candidate entities. The UMLS release version used in this study is 2023AB.
Specifically, we use UMLS \cite{umls} "/search/current" API and set the search term as our entity to retrieve the top 5 Concept Unique Identifiers (CUI) and their names associated with the entity.
For each CUI, we use the "/content/current/CUI/{cui}" API to retrieve the semantic type name for the CUI and then use the semantic\_uri to find the treeid of the semantic type.
Finally, the treeid is mapped to the BioNNE category to finalize the category of an entity.

An entity is finalized as a specific category only when both the language model (either LLM or NER model) and the UMLS heuristics agree on the category.
If UMLS heuristics show that the entity fits multiple categories in the top 5 CUI results, we will use the category type that comes first as the final category.
The mapping between BioNNE category and UMLS semantic type is shown in Table \ref{tab:umls_mapping}.
The mapping is based on the NEREL-BIO paper \cite{nerelbio}, with additional UMLS semantic types added for ANATOMY and LABPROC categories.

\begin{table}
  \caption{BioNNE type and UMLS type mapping}
  \centering
  \begin{tabular}{|c|l|}
    \hline
    BioNNE Category & UMLS semantic types \\
    \hline\hline
    DISO &  B2.2.1.2 Pathologic Function\\
    \hline
    CHEM &  A1.4.1 Chemical\\
    \hline
    ANATOMY & \makecell[l]{A1.2 Anatomical Structure \\ A2.1.5.2 Body Location or Region} \\
    \hline
    LABPROC &  \makecell[l]{B1.3.1.1 Laboratory Procedure \\ B1.3.1.2 Diagnostic Procedure}  \\
    \hline
    INJURY\_POISONING & B2.3 Injury or Poisoning \\
    \hline
    DEVICE & A1.3.1 Medical Device\\
    \hline
    PHYS &  B2.2.1.1 Physiologic Function\\
    \hline
    FINDING &  A2.2 Finding \\
    \hline
  \end{tabular}
  \label{tab:umls_mapping}
\end{table}

\subsection{Acronym Detection}
We use the ScispaCy abbreviation detector \cite{scispacy_ner} to detect acronyms in the abstract.
If the long form of the acronym is identified by the language model and UMLS heuristics as a named entity,
the acronym will be assigned the same category as the long form entity.

\section{Results and Discussion}

Our model achieved 0.348 F1 score on the leaderboard for the BioNNE English track test set.
The F1 score we achieved on the validation set is 0.39, which is close to the test F1 score.
Since the golden answers for the test set have been not released, we will discuss the validation set results here.

The English validation set contains 50 records. The F1 scores for each category sorted in descending order are shown in Table \ref{tab:dev_score}.
Our model performs well on DISO, CHEM, and ANATOMY entities, with F1 scores above 0.5, but the F1 scores for FINDING and PHYS are below 0.3.

\begin{table}
  \caption{F1 Score for each category on the validation set}
  \centering
  \begin{tabular}{|c|c|c|c|}
    \hline
    Category & Precision & Recall & F1 Score \\
    \hline
    DISO & 0.7565 & 0.5613 & 0.6444 \\
    CHEM & 0.7857 & 0.4695 & 0.5878 \\
    ANATOMY & 0.8101 & 0.4082 & 0.5429 \\
    LABPROC & 0.5357 & 0.2632 & 0.3529 \\
    INJURY\_POISONING & 0.2778 & 0.4545 & 0.3448 \\
    DEVICE & 0.6250 & 0.2 & 0.3030 \\
    PHYS & 0.4875 & 0.1444 & 0.2229 \\
    FINDING & 0.1288 & 0.2537 & 0.1709 \\
    \hline
  \end{tabular}
  \label{tab:dev_score}
\end{table}

We examine the model predictions for FINDING and PHYS entities and have the following observations.
\begin{enumerate}
\item The LLM model generates a lot of false positives for the FINDING entities.
The UMLS heuristics can filter out some of the false positives. However, since our UMLS concept search does not search for exact terms, 
some of the LLM-recognized phrases that are part of some FINDING entity word are not filtered out, even though those LLM-recognized phrases (when looked at independently) do not belong to the FINDING type.
For example, for the sentence (in record 27029443\_en), "The authors suggest the algorithm for choosing the order of priority of surgical interventions on coronary and brachiocephalic arteries",
the LLM recognizes "suggest" as a FINDING entity, but the UMLS heuristics do not filter it out because it corresponds to "Abnormal/suggest Ca", which is a FINDING concept.

\item PHYS entity recognition has a low recall. On one hand, PHYS entities sometimes contain generic one-word terms such as "healthy", "lifetime", "size", "shape" and "adults",
which are not recognized by LLM or UMLS as PHYS entities. On the other hand, some long phrases (which are outer nested entities) recognized by LLM as PHYS entities are excluded by UMLS heuristics because the UMLS search
cannot find such terms. For example, "spirometric indicators" and "peripheral blood oxygen saturation" in record 27030325\_en are excluded by UMLS heuristics.
\end{enumerate}

The general takeaways from the results are:
\begin{enumerate}
  \item Context is important in NER task. The same phrases have different entity types in different documents. For example, the words “inflammatory” and “albuminuria” can be PHYS or DISO in different contexts.
Our UMLS heuristics only check the semantic meaning of phrases without looking at the context, which limits the performance of our model.
  \item Our model is often unable to detect the outer nested named entities. For example, it recognizes “cardiac contractility” as PHYS, but fails to recognize “low cardiac contractility” as FINDING. 
Sometimes those outer nested named entities are identified by LLM but rejected by UMLS heuristics because UMLS search cannot find matching terms in the UMLS system.
\end{enumerate}

\subsection{Impact of UMLS Heuristics}
To understand the impact of UMLS heuristics on the system performance, we remove the UMLS heuristics from the system pipeline.
When the LLM model identifies multiple entity types for a phrase, we assign the final entity type according to the
INJURY\_POISONING, ANATOMY, PHYS, DISO, CHEM, LABPROC, DEVICE and FINDING order.
The macro-F1 score on the validation set without UMLS heuristics is 0.2151, which is significantly lower than the F1 score (0.348) when UMLS heuristics are used.
The F1 scores for every category are shown in Table \ref{tab:dev_no_umls_score}.
We observe that without UMLS heuristics, the precision score for each category becomes lower. This is because the LLM model generates many false positives.
In terms of recall, only the INJURY\_POISONING and ANATOMY types have better recall scores when UMLS heuristics are not used. 
This is because we prioritize assigning the final entity type as INJURY\_POISONING and ANATOMY when the LLM detects multiple matching entity types.
This result shows that UMLS heuristics play an important role in eliminating the false positives of LLM predictions.

\begin{table}
  \caption{F1 Score for each category on the validation set when UMLS heuristics are not used}
  \centering
  \begin{tabular}{|c|c|c|c|}
    \hline
    Category & Precision & Recall & F1 Score \\
    \hline
    DISO & 0.4551 & 0.4676 & 0.4613 \\
    CHEM & 0.3731 & 0.3049 & 0.3356 \\
    ANATOMY & 0.2036 & 0.4629 & 0.2828 \\
    LABPROC & 0.1964 & 0.0973 & 0.1302 \\
    INJURY\_POISONING & 0.1304 & 0.5455 & 0.2105 \\
    DEVICE & 0.069 & 0.08 & 0.0741 \\
    PHYS & 0.0742 & 0.10 & 0.0852 \\
    FINDING & 0.1357 & 0.1463 & 0.1408 \\
    \hline
  \end{tabular}
  \label{tab:dev_no_umls_score}
\end{table}

\subsection{Training Dataset and Results}
The Mixtral 8x7B model officially supports five languages, which do not include Russian.
Therefore, we only attempt the English track.
It is possible that other LLM models that support Russian can be used to accomplish the Russian/bilingual track using the same modeling pipeline.
The method we use in this study is few-shot prompt engineering and we only use two examples from the training dataset to construct the LLM prompt.
We did not include more examples in the prompt, as the input token size of LLM is limited and the cost of processing a long prompt is high. 
We also do not expect adding more examples in the prompt would improve the performance significantly, as the prompt is mostly used to guide
the model to generate outputs in the desired format. The entities recognized by the model are mostly based on the original knowledge of the model.
However, we believe the remaining training examples will be useful if we want to finetune the LLM model on the BioNNE task in the future.

\section{Future Work}
There are two directions for future improvement.

First, we can improve the heuristics for determining the category of an entity. Our current algorithm is weak at detecting the outer nested entities. Even though the LLM model
recognizes them, we fail to assign the correct category for those entities using the existing UMLS heuristics.
We can add a new set of heuristics for detecting the outer nested entities. Specifically, if a long phrase contains an inner entity that is a named entity,
we can assign the outer entity with the same category as the inner entity or assign the outer entity with a category detected by the LLM.

Second, we can improve the performance of the LLM model. Currently LLM model generates many false positives and it may assign several entity types to a single phrase.
For example, it recognizes "neuronal dysfunction" as DISO, ANATOMY, PHYS, FINDING and CHEM in the sentence 
"the severity of coronary artery lesions and low cardiac contractility affect the degree of cerebral ischemia and neuronal dysfunction detected by spectral EEG power" (in record 25726786\_en).
This is probably because the general-purpose LLM model (Mixtral 8x7B instruct) trained on the general text does not have the domain knowledge of the biomedical field. 
Therefore it cannot distinguish the subtle differences among the BioNNE entity types.
We can fine-tune the LLM model on the BioNNE training data as well as other biomedical NER datasets to improve the performance of the model in this task.

Although the Mixtral 8x7B model does not officially support the Russian language,
a recent study by Fenogenova et al. \cite{fenogenova2024} shows that Mistral 7B (a smaller version of Mixtral 8x7B)
performs well on Russian language benchmarks compared to other LLMs.
Future work can explore using Mixtral or other LLM models to solve the BioNNE bilingual NER problems using prompt engineering techniques in the paper or fine-tuning the model with
bilingual datasets.

\section{Conclusions}
We build a system that uses a general-purpose LLM, a biomedical domain NER model and UMLS-based heuristics to extract nested named entities from biomedical text.
Our model achieved an F1 score of 0.39 and 0.348 on the BioNNE English validation and test sets. Although the results are not comparable to those of the state-of-the-art models \cite{nerelbio}, that
were trained on the domain-specific dataset, our study demonstrates the potential of using general-purpose LLM and prompt engineering with domain-specific rules to solve biomedical NER tasks.

\section*{Acknowledgements}

We want to thank the Data Science @ Georgia Tech (DS@GT) CLEF team for their support and acknowledge the use of Grammarly \cite{grammarly} to proofread this paper.

\bibliography{main}

\end{document}